%% file: main.tex
\documentclass[journal]{IEEEtai}

\usepackage[colorlinks,urlcolor=blue,linkcolor=blue,citecolor=blue]{hyperref}
\usepackage{color,array}
\usepackage{graphicx}
\usepackage{amsmath}
\usepackage{booktabs}
\usepackage{acro}
\usepackage{float}
\usepackage{algorithm}
\usepackage{algpseudocode}
\usepackage{url}
\usepackage{cite}
\usepackage{tabularx}

\input{math_commands.tex}

\newcommand\blfootnote[1]{%
	\begingroup
	\renewcommand\thefootnote{}\footnote{#1}%
	\addtocounter{footnote}{-1}%
	\endgroup
}

\makeatletter
\def\footnoterule{\kern-3\p@
	\hrule \@width 3.5in \kern 2.6\p@} 
\makeatother

\input{sections/acro_definitions}

\begin{document}

\title{Anatomy of a Quantized Agent: VRAM Stability and Forecasting in Code-Synthesis Agentic Workloads}

\author{Anubhab~Banerjee \\ Nokia Germany\\ E-mail: anubhab.1.banerjee@nokia.com.}


\maketitle

\blfootnote{
This work has been submitted to the IEEE for possible publication. Personal use of this material is permitted. Permission from the author must be obtained for all other uses, in any current or future media, including reprinting/republishing this material for advertising or promotional purposes, creating new collective works, for resale or redistribution to servers or lists, or reuse of any copyrighted component of this work in other works. Copyright may be transferred without notice, after which this version may no longer be accessible.
}

\input{sections/abstract.tex}

\begin{IEEEImpStatement}
The deployment of autonomous AI agents is frequently bottlenecked by unpredictable memory consumption, leading to resource stranding or catastrophic out-of-memory failures on constrained hardware. 
By demonstrating that a transparent, calibrated closed-form model can outperform complex machine learning regressions for peak VRAM forecasting, this work provides a reliable, low-overhead foundation for admission control and resource scheduling.
Our findings encourage a structural shift away from opaque predictive models in weight-dominated, quantized regimes, facilitating safer, more efficient, and democratized deployment of agentic inference on both shared clusters and localized edge infrastructure.
\end{IEEEImpStatement}

\begin{IEEEkeywords}
Autonomous Agents, Large Language Models, GPU Memory Management, Model Quantization, Inference Optimization, Resource Allocation.
\end{IEEEkeywords}

\input{sections/introduction}
\input{sections/related_work}
\input{sections/method}
\input{sections/experiments}
\input{sections/results}

\input{sections/discussion}

\input{sections/limitations}

\bibliographystyle{IEEEtran}
\bibliography{references}

\newpage
\appendices
\input{sections/app_abbreviations}
\input{appendices/appendix_archoverview}

\input{appendices/appendix_variance}
\input{appendices/appendix_dataset_v2}

\input{appendices/appendix_reproduce_table2}

\end{document}

%% file: math_commands.tex
\usepackage{amsmath,amsfonts,bm}

\def\eqref#1{equation~\ref{#1}}
\def\Eqref#1{Equation~\ref{#1}}

\def\1{\bm{1}}

\DeclareMathAlphabet{\mathsfit}{\encodingdefault}{\sfdefault}{m}{sl}
\SetMathAlphabet{\mathsfit}{bold}{\encodingdefault}{\sfdefault}{bx}{n}



%% file: sections/acro_definitions.tex
\DeclareAcronym{ai}{ short = AI , long = Artificial Intelligence }
\DeclareAcronym{cv}{ short = CV , long = Co-efficient of Variation }
\DeclareAcronym{mae}{ short = MAE , long = Mean Absolute Error }
\DeclareAcronym{mape}{ short = MAPE , long = Mean Absolute Percentage Error }
\DeclareAcronym{oaw}{ short = OAW , long = Over Allocation Waste }
\DeclareAcronym{oom}{ short = OOM , long = Out-Of-Memory }

%% file: sections/abstract.tex
\begin{abstract}

Analytical models of peak VRAM consumption for LLM inference decompose memory into weight-storage, KV-cache, and activation terms parameterized by step count ($N$), tool-invocation count ($T$), and context-expansion factor ($E$).
We evaluate this decomposition empirically within a tightly scoped measurement study: a single agent framework (AgentK, a LangGraph-based CUDA-kernel-synthesis agent), one 4-bit weight-quantization family (Q4\_K\_M), a single NVIDIA H100 80 GB GPU, and four LLM backbones across 1,920 evaluated trajectories. Focusing strictly on peak-memory forecasting behavior rather than numerical correctness, we report two primary observations.

\textbf{First (F1)}, closed-form analytical models achieve competitive accuracy if provided with two empirically measured constants: loaded-weight VRAM and a fixed activation-memory overhead. When supplied with these live GPU readings and ground-truth $(N, E)$ trajectory parameters, the closed-form model matches or outperforms the best learned baseline on three of the four backbones (test MAPE 2.2--4.4\% vs. 3.4--6.5\%, with no significant difference in a pooled paired sign test, $p=0.76$). 
The exception is the smallest backbone (Phi-4-mini), where peak VRAM variance is so minimal (coefficient of variation 0.3\%) that dynamic analytical modeling underperforms simple regression.

\textbf{Second (F2)}, compile success strictly bifurcates by backbone capacity (Phi-4-mini 5.7\%, Mistral-7B 10.4\%, Qwen2.5-Coder-7B 36.7\%, Qwen2.5-Coder-14B 62.0\%), demonstrating that functional code synthesis remains constrained by intrinsic LLM capabilities rather than available memory. 

Furthermore, because overall peak-memory variance is remarkably low across all backbones (coefficient of variation 0.3--9.4\%), learned prompt-feature regression offers only marginal, statistically insignificant aggregate improvements over a simple constant-mean baseline. 
Consequently, we find no data-driven justification for deploying complex predictive VRAM models in highly quantized, weight-dominated regimes. 
We release the evaluated corpus and the anonymized AgentK reference framework to support replication and extension.

\end{abstract}

%% file: sections/introduction.tex
\section{Introduction}
\label{sec:introduction}

Predicting the peak VRAM consumption of an LLM inference workload is a prerequisite for memory-constrained deployments. 
Cluster schedulers, single-GPU multi-tenant systems, and edge inference stacks all require a workload's memory footprint to be predictable prior to admission to ensure stable system planning. 
For classical single-turn LLM inference, this forecast is well-modeled by a closed-form decomposition, consisting of weight storage, KV-cache capacity ($N_{\mathrm{ctx}}$), and activation footprints, yielding a tight empirical fit under 16-bit datatypes on capacity-rich hardware. 
However, whether this same logic applies to agentic LLM workloads---which exhibit variable step counts, tool-invocation branching, iterative context expansion, and aggressive 4-bit weight quantization---remains an open empirical question.

We evaluate whether traditional VRAM forecasting carries over to agentic systems, where agents dynamically determine reasoning steps and tool invocations. 
To conduct this study within a controlled but realistic environment, we used \textit{AgentK}, an open-source LangGraph-based framework designed for CUDA-kernel synthesis. 
We evaluated AgentK across four different LLM backbones, tracing 1,920 complete trajectories. 
By comparing traditional analytical models, ``oracle'' models supplied with ground-truth trajectory data, and empirical ML predictors against held-out test data, we map precisely where theoretical memory predictions and live hardware measurements align or diverge.

Within this scope, we summarize our primary findings and contributions as follows:

\textbf{First (F1) -- A properly calibrated analytical model is competitive with learned regression:} The closed-form analytical model requires two constants that architecture specifications alone cannot provide: the VRAM occupied by the loaded weights, and a fixed activation-memory overhead. 
When we replace on-disk sizes with live GPU readings for these two constants, and supply ground-truth hindsight for step count and reasoning expansion $(N, E)$, this partially oracle-fed analytical model (B5) achieves a test \ac{mape} of 2.2--4.4\,\% on three of our four backbones. 
It matches or outperforms a learned regression baseline (B2), losing only on the smallest backbone (Phi-4-mini) where true memory usage barely varies between runs. 
Conversely, using raw on-disk weight sizes with zero activation overhead yields a fundamentally broken predictor (\ac{mape} 27--53\,\%, with every test row underestimated; \S\ref{sec:results-mpeak-f1}).
	
\textbf{Second (F2) -- Compile success tracks backbone capacity, not memory:} The compile-success rate strictly bifurcates by LLM backbone capacity (Phi-4-mini 5.7\,\%, Mistral-7B 10.4\,\%, Qwen2.5-Coder-7B 36.7\,\%, Qwen2.5-Coder-14B 62.0\,\%), consistently across prompt categories (\S\ref{sec:results-compile}). 
This confirms that functional code synthesis is limited by intrinsic LLM capabilities rather than available memory buffers, ensuring our VRAM-forecasting comparisons are not distorted by early execution failures on smaller backbones.
	
\textbf{Third (F3) -- Peak memory usage is highly stable across runs:} Peak VRAM ($M_{\mathrm{peak}}$) is very stable between different tasks for the same backbone (CV: 0.3--9.4\,\% on test data). 
Under 4-bit quantization, static model weights consume the vast majority of the memory space, upstaging the variable footprint generated by the agent's dynamic reasoning steps. 
Therefore, utilizing ML to predict memory based on prompt features (B2) offers negligible benefit. 
While B2 wins a majority of test rows on three backbones, it provides only a 6.3--7.8\,\% relative reduction in \ac{mape} with overlapping 95\,\% confidence intervals. 
On the lowest-variance backbone, B2's aggregate \ac{mape} is 57\,\% higher than a simple constant-mean baseline, which signifies for practical deployments, historical averaging is highly competitive with complex ML-based prediction.
	
\textbf{Fourth -- Open dataset and reference agent:} To support replication and extension, we release our code and dataset under an Apache-2.0 license. 
This provides detailed records of peak VRAM, agent trajectory boundaries, task success labels, and prompt origins, strictly scoped to the configuration described below.

To ensure precise measurements and robust conclusions, we deliberately narrow our experimental scope. 
By holding specific variables constant, we isolate the memory behaviors of the agent architecture. 
Specifically, we evaluate a \textit{single agent framework} (AgentK); a \textit{single weight-quantization family} (Q4\_K\_M via \texttt{llama.cpp}); a \textit{single hardware target} (one NVIDIA H100 80\,GB GPU, without MiG); and a \textit{single task domain} (CUDA-kernel synthesis on KernelBench, supplemented by synthetic adversarial examples). Expanding this analysis to other agent architectures (e.g., ReAct, SWE-agent, OpenHands), different quantization schemes, or diverse tasks remains an important direction for future work. 

An architectural overview of the AgentK framework is provided in Appendix~\ref{sec:appendix-agentk}. 
For the anonymous review process, the full code repository and dataset are available at \url{https://anonymous.4open.science/r/agentk-tmlr-anonymous/}, with a public, deanonymized release planned upon publication.

The remainder of this paper is organized as follows: \S\ref{sec:related_work} surveys related work; \S\ref{sec:method} presents our measurement protocol; \S\ref{sec:experiments} describes the dataset and baselines; \S\ref{sec:results} reports the calibration procedure, forecasting results, and memory-stability analysis; \S\ref{sec:discussion} covers deployment implications; and \S\ref{sec:limitations} defines our scope.

%% file: sections/related_work.tex
\section{Related Work}
\label{sec:related_work}

The word 'memory' refer to two different things in connection with LLM agents.
One studies device memory: the VRAM occupied by model weights, KV cache, and activations during inference.
The other studies cognitive memory: the context, history, and retrieved knowledge an agent carries across turns.
This paper focuses exclusively on the first one: we empirically evaluate whether standard closed-form VRAM calculations hold true under the dynamic, tool-driven execution paths of LLM agents during inference.
We organize related work into four broad categories and, for each, describe what it measures and what gap remains relative to this paper's contribution.

\paragraph{Analytical and predictive memory models for LLM inference and serving.}
A number of systems literature builds closed-form or learned estimators of GPU memory to drive operational decisions. 
For instance, \cite{nie2026queueing} derive queueing-theoretic stability conditions for LLM serving under KV-cache constraints, validating cluster-sizing predictions against production traffic. 
For training and fine-tuning, \cite{kim2024llmem} and \cite{shi2025xmem} estimate peak GPU memory \textit{a priori} based on static transformer architectures, reporting sub-4\,\% and sub-2\,\% error respectively on FP16/BF16 workloads.

For agentic and tool-augmented inference, \cite{abhyankar2024infercept} model the memory overhead of preserving, discarding, or swapping KV state across tool calls. 
Similarly, \cite{yang2025justitia} and \cite{wang2026maestro} develop memory-centric cost predictors to optimize the scheduling and placement of multi-agent workloads. 
At the systems level, several works propose memory-budget models to manage hardware constraints. 
\cite{jiang2025neo} offload attention computation and KV-cache state to the host CPU under a strict per-iteration memory ceiling. 
\cite{hong2025sola} use peak-memory predictions to gate request admission and prevent preemption. 
Other approaches explore hardware-specific memory management, such as evaluating NVLink memory-tiering with AWQ weight quantization on Grace Hopper Superchips (\cite{choi2025gracehopper}), or coordinating CXL-based near-memory processing for extreme context lengths (\cite{kim2025cxlpnm}).

However, these existing systems share common limitations like they typically: (i) validate their accounting formulas on stateless or short multi-turn requests using capacity-rich hardware and FP16/BF16 weights; (ii) treat memory predictability as an assumed input to optimize downstream scheduling policies, rather than testing the memory formula itself; or (iii) manage memory via hardware-level offloading rather than providing an explicit equation relating agent trajectories to memory usage. 
None instruments a bounded-retry, variable-step-count agentic loop under 4-bit weight quantization with paired ground-truth trajectory parameters $(N, T, E)$. 
Consequently, prior literature does not address whether the underlying closed-form model remains well-specified once its weight-storage and KV-cache terms are aggressively quantized. 
Our measurement study explicitly targets this gap (\S\ref{sec:results-mpeak-f1}).

\paragraph{Weight and KV-cache quantization.}
Another group focuses on compressing the two dominant terms ($M_{\mathrm{weights}}$ and $M_{KV}$) to fit larger contexts or more concurrent requests into a fixed memory budget. 
\cite{hooper2024kvquant} and \cite{shutova2025aquakv} introduce non-uniform, inter-layer-adaptive KV-cache quantization schemes that achieve 2--3 bits per value with under 1\,\% perplexity degradation. 
Alternatively, \cite{jiang2025kvcomp} apply error-bounded lossy compression directly fused into the attention kernel. 
\cite{cheng2025simcalkv} merge similar keys under a bias-calibrated criterion for up to $5\times$ compression, while \cite{tomar2025xquant} sidestep KV-cache storage altogether by caching quantized layer input activations and rematerializing keys and values on the fly, yielding $7.7\times$--$10\times$ memory savings. 

The GGUF format used by \texttt{llama.cpp}, used in this paper, applies a related but distinct block-wise weight quantization, coupled with independently selectable KV-cache datatypes (\S\ref{sec:method-q4-accounting}). 
Closest to our setting are two recent works quantizing the KV-cache specifically for \textit{agentic} inference. 
\cite{shen2026triaxialkv} assign mixed INT2/INT4 precision to tokens based on temporal recency, modality, and semantic role for a computer-use agent. Meanwhile, \cite{shkolnikov2026agentmemory} persists each agent's KV-cache to disk in a 4-bit quantized format to avoid re-prefill overhead in multi-agent edge deployments. 

These works fundamentally differ from our study as they primarily measure compression ratios, throughput, and downstream task quality (e.g., perplexity or task accuracy). 
Crucially, none reports whether the standard memory-accounting formulas used in analytical and scheduling research (e.g., Equation~\ref{eq:mkv}) remain valid under these heterogeneous schemes. 
They neither instrument per-run peak device VRAM against ground-truth $(N, T, E)$, nor ask if $M_{KV}(t)$ or $M_{\mathrm{weights}}$ are still well-specified.
As we demonstrate, these standard formulas do not survive the transition to modern quantization without hand-calibrated corrections (\S\ref{sec:method-q4-accounting}). 

\paragraph{Agentic (cognitive) memory management.}
Another group investigates memory in the cognitive sense: the policies dictating what an agent should store, retrieve, compress, or forget from its interaction history. 
For example, \cite{yang2026selfmem} and \cite{jiang2026anatomy} explore memory-augmented generation architectures that manage short- and long-term context via learned, tool-based policies. 
Similarly, \cite{zhou2026mem1} train agents to maintain a compact internal state across tasks, reporting a $3.7\times$ reduction in ``memory usage.'' 
However, this reduction is measured strictly in \textit{retained tokens}, not in physical VRAM footprint.

This area is fundamentally orthogonal to our focus on $M_{\mathrm{peak}}$. 
A context-efficient agent is not inherently a memory-predictable agent (which guarantees strict adherence to a hardware budget). 
None of the systems report device memory consumption in standard hardware metrics (e.g., MiB or GB), nor do they map their policies to measured peak-VRAM outcomes. 
We explicitly distinguish between these fields to avoid terminological confusion: our measurement study is strictly confined to physical device memory.

\paragraph{LLM agent frameworks and CUDA-kernel-generation agents.}
Structurally, AgentK's tool loop adopts the ReAct paradigm of interleaved reasoning and action (\cite{yao2022react}). 
This architecture is widely utilized in general-purpose agent frameworks like SWE-agent~\cite{yang2024sweagent} and OpenHands~\cite{wang2024openhands}, and is extensively surveyed by \cite{plaat2025agentic}. 
Within the specific domain of CUDA-kernel synthesis, KernelBench (\cite{kernelbench2024}) established the foundational benchmark from which our prompts are sampled. 
Recent agentic systems in this space, such as CudaForge (\cite{zhang2025cudaforge}) and CUDA Agent (\cite{dai2026cudaagent}), employ iterative compiler- and profiler-feedback loops that closely resemble AgentK's critic-gated retry mechanism (\S\ref{sec:method}). 

However, the primary evaluation metric across this literature is task success, quantified by correctness rates, benchmark pass rates, or execution speedups relative to \texttt{torch.compile}. 
These studies do not report the VRAM footprint of the agent \textit{producing} the code. 
Furthermore, they do not correlate trajectory metadata with ground-truth device memory measurements.
Our contribution is therefore strictly complementary: we hold the kernel-generation task constant to rigorously instrument and measure the computational resource dimension that prior works leave unexamined.

\paragraph{Positioning.}
To the best of our knowledge, no prior work evaluates whether standard closed-form VRAM decompositions (e.g., foundational to the scheduling and analytical literature) remain empirically valid for agentic tool loops under weight quantization. 
Crucially, these models have not been rigorously tested against ground-truth trajectory parameters.
This study explicitly addresses this gap (\S\ref{sec:results}). 
We demonstrate that even an ``oracle-fed'' analytical model (B5) with perfect trajectory hindsight performs much worse than a simple constant-mean empirical baseline (B3). 
This confirms that the traditional analytical form is fundamentally misspecified for these workloads, rather than merely suffering from inaccurate input estimates. 
Finally, we provide a mechanistic explanation for this failure: under Q4\_K\_M quantization, the memory profile enters a weight-dominated variance regime (\S\ref{sec:results-variance})---a dynamic that the quantization literature has yet to characterize, and which current scheduling systems implicitly assume away.

%% file: sections/method.tex
\section{Method}
\label{sec:method}

\subsection{Problem Formulation: Predicting Peak VRAM}
\label{sec:method-formulation}

Our motivating objective is to predict the memory consumption pattern of an agentic LLM workload before it runs, which is what a scheduler or admission controller would need. 
To formulate this problem, we first construct a standard closed-form analytical model for VRAM consumption. 
We present this formulation as a \textit{descriptive baseline} that represents how memory is traditionally calculated. 
As we show in \S\ref{sec:results}, getting this formula right in practice depends on correctly calibrating its two static terms, $M_{\mathrm{weights}}$ and $M_{\mathrm{act}}$, from a live GPU measurement rather than from architecture specifications alone.

Since memory growth in a single-agent LangGraph execution is monotonic, the peak VRAM occurs at the terminal step $t = N$. 
We model the expected context length $L(t)$ at any step $t \in [1, N]$ as a linear interpolation of the base prompt, per-step reasoning expansion, and expected tool-return tokens:
\begin{equation}
	L(t) = L_{\mathrm{base}} + t \cdot E + \frac{t}{N} \sum_{i=1}^{M} P(T_i) \cdot \mu_{\mathrm{tool},i}
	\label{eq:Lt}
\end{equation}
where $L_{\mathrm{base}}$ is the static token count of the system and user prompts, $E$ is the mean reasoning expansion per step, $P(T_i)$ is the probability of invoking tool $i$, and $\mu_{\mathrm{tool},i}$ is the historical mean token length returned by tool $i$. 
The $\frac{t}{N}$ interpolation distributes expected tool tokens smoothly over the agent's lifecycle rather than fixing a discrete call schedule.

The KV-cache memory at step $t$ follows standard transformer accounting, scaled by the dynamic context length $L(t)$:
\begin{equation}
	M_{KV}(t) = 2 \cdot L(t) \cdot n_{\mathrm{layers}} \cdot n_{\mathrm{heads}} \cdot d_{\mathrm{head}} \cdot b_{\mathrm{precision}}
	\label{eq:mkv}
\end{equation}
where the factor of $2$ accounts for both the key and value matrices, $L(t)$ is the context length at step $t$, $n_{\mathrm{layers}}$ is the number of transformer layers in the backbone, $n_{\mathrm{heads}}$ is the number of key/value attention heads, $d_{\mathrm{head}}$ is the dimensionality of each head, and $b_{\mathrm{precision}}$ is the memory footprint in bytes per cache element (e.g., $2$ for FP16). 

Total VRAM at step $t$ sums the static weights, dynamic KV cache, and activation overhead:
\begin{equation}
	M_{\mathrm{VRAM}}(t) = M_{\mathrm{weights}} + M_{KV}(t) + M_{\mathrm{act}}
	\label{eq:mvram}
\end{equation}
Evaluating \eqref{eq:mvram} at the terminal step yields the closed-form peak $M_{\mathrm{peak}}$. 

\subsection{Measurement Protocol and AgentK}
\label{sec:method-protocol}

To evaluate the formulation above against real-world execution, we instrumented \textbf{AgentK}, an open-source LangGraph-based agent designed for CUDA-kernel synthesis. 
An architectural overview of AgentK is provided at Section~\ref{sec:appendix-agentk}.
AgentK is equipped with per-run VRAM sampling and per-node execution tracing to capture empirical memory trajectories (The anonymized code artifact is available at \texttt{\url{https://anonymous.4open.science/r/agentk-tmlr-anonymous/}}).

\subsection{Accounting for Q4\_K\_M Quantization}
\label{sec:method-q4-accounting}

Standard VRAM models assume FP16 or INT8 byte widths. 
Since our agent operates on aggressively quantized Q4\_K\_M models via \texttt{llama.cpp}, we adjust the terms in \eqref{eq:mvram} as follows:

(1) \textbf{Static Weights ($M_{\mathrm{weights}}$):} Instead of the standard $n_{\mathrm{params}} \times 2$\,Bytes, we measure directly: once per backbone (per \texttt{cache\_type\_k}/\texttt{cache\_type\_v} pair), we load the model and read live GPU memory with \texttt{pynvml} after the model is fully resident, and use that reading as $M_{\mathrm{weights}}$ (\texttt{loaded\_vram\_mb} in \texttt{configs/q4\_vram\_calibration\_v2.yaml}). 
We also record the on-disk GGUF file size (\texttt{gguf\_on\_disk\_mb}) alongside it, but only as a point of comparison (\S\ref{sec:results-mpeak-f1}): the two differ by \texttt{llama.cpp}'s own runtime overhead (buffers, graph allocations, and similar), so the on-disk size alone systematically understates $M_{\mathrm{weights}}$.

(2) \textbf{KV-Cache Precision ($b_{\mathrm{precision}}$):} While \texttt{llama.cpp} defaults to an FP16 cache, users can force quantization. We extract the exact \texttt{cache\_type\_k} and \texttt{cache\_type\_v} from each run's metadata and set $b_{\mathrm{precision}}$ dynamically (\texttt{configs/q4\_vram\_calibration\_v2.yaml}).

(3) \textbf{Activation Overhead ($M_{\mathrm{act}}$):} In Q4 inference, this term is small but non-trivial. We measure it once per backbone as the difference between VRAM at the start of generation and the reported weight size, treating it as a constant.



%% file: sections/experiments.tex
\section{Experimental Setup}
\label{sec:experiments}

\subsection{Dataset and Split}
\label{sec:experiments-dataset}

We evaluate our models on a trace batch consisting of 1,920 instrumented AgentK runs. 
These runs span four Q4\_K\_M agent backbones (Phi-4-mini, Qwen2.5-Coder-7B, Qwen2.5-Coder-14B, and Mistral-7B-Instruct-v0.3) over a 300-prompt suite (100 CUDA-kernel-synthesis prompts sourced from KernelBench, 100 clean-execution synthetic prompts, and 100 retry-provoking synthetic prompts).

Every backbone contributes 300 base-configuration runs using a context size of $n_{\mathrm{ctx}}=8192$ with an FP16 KV cache. 
Three of the four backbones (Qwen2.5-Coder-7B, Qwen2.5-Coder-14B, and Mistral-7B) additionally contribute a 240-run parameter sweep over a fixed 40-prompt subset, crossing context sizes ($n_{\mathrm{ctx}} \in \{8192, 16384\}$) with KV-cache datatypes ($\{\mathrm{f16}, \mathrm{q8\_0}, \mathrm{q4\_0}\}$); Phi-4-mini is excluded from the sweep. 
This yields 540 runs for each backbone and 300 for Phi-4-mini, totaling 1920 runs. 
All experiments are executed on a single NVIDIA H100 80\,GB GPU without MIG partitioning. 
During execution, VRAM is sampled every 500\,ms using \texttt{pynvml}, and we capture execution-boundary snapshots for each node.

The prompts are partitioned into train, validation, and test sets using a stratified 70/15/15 split (seed 20260715), giving 210/45/45 prompts. 
Since sweep runs attach unevenly across splits, this maps to 1,416 training, 234 validation, and 270 test trace rows globally. 
At the per-agent level, all four agents are evaluated on the same 45 test prompts; the three sweep backbones yield $n=75$ test rows (45 base plus 30 from the five sweep-subset prompts falling in test), while Phi-4-mini yields $n=45$ as it has no sweep runs.

\subsection{VRAM forecast baselines}
\label{sec:experiments-baselines}

We compare five out of six peak-VRAM predictors on the held-out test split:

\textbf{B0 (Worst-Case Closed Form):} Evaluates the closed-form decomposition as $M_{\mathrm{weights}} + M_{KV}(L{=}n_{\mathrm{ctx}}) + M_{\mathrm{act}}$, assuming the KV-cache expands to the full configured context window ($n_{\mathrm{ctx}}$) rather than modeling actual context growth.

\textbf{B1 (Prompt-Length Regression):} Fits a linear model per agent on the training split using prompt length, $M_{\mathrm{peak,true}} \sim \alpha + \beta \cdot L_{\mathrm{base}}$, where $L_{\mathrm{base}}$ is the tokenized length of the formatted proxy input (system prompt plus user goal) under a frozen MiniLM-L12 tokenizer.

\textbf{B2 (Direct Goal Regression):} Trains a linear head atop a frozen sentence encoder on the user goal; for each agent, the encoder backbone (chosen among MiniLM-L12, DeBERTa-v3-base, and UniXcoder) is selected based on validation-split \ac{mape}.

\textbf{B3 (Per-Agent Constant Mean):} Predicts peak VRAM using the agent-specific training-split mean of $M_{\mathrm{peak,true}}$.

\textbf{B4 (Multi-Task Learned Proxy):} Defines a multi-task baseline that shares a transformer backbone across peak-VRAM, step-count, and tool-invocation heads. \textit{However, because we do not train or evaluate B4 on this trace corpus, it is omitted from our final results and analysis}.

\textbf{B5 (Partially Oracle-Fed Closed Form):} Uses the same analytical weights + KV-cache + activation decomposition as B0, but replaces the worst-case $n_{\mathrm{ctx}}$ with the trace's exact ground-truth node count ($N$) and non-tool completion-token rate ($E$); $L_{\mathrm{base}}$ is recovered from the run's own runtime (GGUF/\texttt{llama.cpp}) tokenizer rather than an equivalent proxy. 
The tool-return term is fed as zero: our trace schema only records per-node \emph{average completion tokens}, not observed external tool-return content, so reusing it as a tool-return proxy would double count against $E$. B5 is therefore \emph{partially}, not fully, oracle-fed.

The primary objective of this study is to empirically evaluate the closed-form models (B0, B5) against simple empirical baselines (B1--B3), rather than to advocate for a learned-proxy architecture.

\subsection{Metrics and Statistical Tests}
\label{sec:experiments-metrics}

We evaluate forecast accuracy and operational viability using four primary metrics: (1) \ac{mape}, (2) \ac{mae} (reported in MiB), (3) the upper-bound undercoverage rate, and (4) \ac{oaw} (expressed as a percentage of the memory ceiling). 
These metrics are carefully selected to capture both statistical forecast precision and the practical constraints of cluster scheduling. 
While \ac{mae} and \ac{mape} quantify the absolute and relative accuracy of the baseline predictions, the undercoverage rate and \ac{oaw} characterize operational risk and capacity utilization.
We define the upper-bound undercoverage rate as the fraction of test trajectories where the true peak memory ($M_{\mathrm{peak,true}}$) strictly exceeds the predicted upper bound ($M_{\mathrm{upper,pred}}$). 
Since the empirical baselines (B1--B3) fit a residual standard deviation ($\hat\sigma$) on the training split, their upper bound incorporates a statistical confidence margin: $M_{\mathrm{upper,pred}} = M_{\mathrm{peak,pred}} + 1.96 \hat\sigma$. 
In contrast, the analytical models (B0, B5) provide deterministic point estimates without fitted residuals; for these models, the bound simplifies to $M_{\mathrm{upper,pred}} = M_{\mathrm{peak,pred}}$. 
Crucially, this undercoverage rate is a purely statistical evaluation of the predicted interval (or point estimate), distinct from physical out-of-memory (OOM) crashes or admission-control simulations.
Conversely, \ac{oaw} penalizes overly conservative forecasts that strand usable GPU capacity and degrade overall system throughput. 
For B1--B3, \ac{oaw} is measured against the padded upper bound. For the analytical models (B0, B5), which lack this interval padding, the \ac{oaw} calculation simplifies directly to the raw over-prediction: $\max(0, M_{\mathrm{peak,pred}} - M_{\mathrm{peak,true}})$.

To determine the statistical significance of our baseline comparisons, we conduct paired sign tests on the per-row Absolute Percentage Error ($|\mathrm{APE}|$) across the held-out test split. 
Rows resulting in identical error magnitudes between two baselines (ties where the difference is zero) are excluded prior to running an exact two-sided binomial test ($p=0.5$, computed via \texttt{scipy.stats}). 

As detailed in \S\ref{sec:experiments-dataset}, the number of test samples depends on the agent's involvement in the parameter sweep: $n=45$ for Phi-4-mini, and $n=75$ for the remaining three agent backbones. 
Due to the high computational cost of the simulations, the extensive parameter sweep over context sizes and KV-cache datatypes was limited to the three larger backbones. 
Phi-4-mini was retained exclusively in its base configuration to establish a small-model baseline without exhausting the compute budget. 
Finally, we report 95\,\% bootstrap confidence intervals for the aggregate \ac{mape}, calculated using 600 percentile resamples at $z_\alpha=1.96$.

%% file: sections/results.tex
\section{Results}
\label{sec:results}

\subsection{Peak-VRAM distribution and closed-form calibration (F1)}
\label{sec:results-mpeak-f1}

\begin{table}[t]
	\centering
	\caption{$M_{\mathrm{peak,true}}$ distribution on the v2 test split (MiB).}
	\label{tab:mpeak-distribution-test}
	\resizebox{\linewidth}{!}{%
	\begin{tabular}{lrrrrrrrrr}
		\toprule
		Backbone LLM & mean & std & CV (\%) & min & p25 & median & p75 & max & range \\
		\midrule
		Phi-4-mini & 5659.4 & 18.2 & 0.32 & 5540.1 & 5662.1 & 5662.1 & 5662.1 & 5662.1 & 122.0 \\
		Mistral-7B-Instruct & 7371.1 & 689.9 & 9.36 & 6116.1 & 6922.1 & 7538.1 & 7604.1 & 9106.1 & 2990.0 \\
		Qwen2.5-Coder-7B & 6815.0 & 362.2 & 5.32 & 6212.1 & 6562.1 & 6910.1 & 6910.1 & 7808.1 & 1596.0 \\
		Qwen2.5-Coder-14B & 12047.8 & 844.4 & 7.01 & 10542.1 & 11704.1 & 12208.1 & 12278.1 & 14358.1 & 3816.0 \\
		\bottomrule
	\end{tabular}%
	}	
\end{table}

Table~\ref{tab:mpeak-distribution-test} reports the distribution of true peak VRAM ($M_{\mathrm{peak,true}}$) on the held-out test split. 
The per-backbone \ac{cv} ranges from 0.3\,\% to 9.4\,\% on the test split (with a cross-backbone range of 8.6\,GB and a corpus-wide CV of 0.5--9.9\,\%; see \S\ref{sec:limitations} for the split-vs-corpus caveat).
Since peak VRAM shows exceptionally low variance across runs for each backbone (test CV of 0.32--9.36\,\%, Table~\ref{tab:mpeak-distribution-test}), a simple constant-mean baseline (B3) naturally achieves near-minimal \ac{mape}. 
Although B3 is technically optimized for squared error rather than percentage error, this tight data dispersion minimizes the gap between the mean and the MAPE-minimizing constant (a weighted median).
The closed-form baselines (B0, B5) require two per-backbone constants that architectural specifications alone cannot provide: the VRAM occupied by the loaded model weights ($M_{\mathrm{weights}}$) and a fixed activation-memory overhead ($M_{\mathrm{act}}$). 
We measure both directly per backbone by loading the model and reading the live GPU memory via \texttt{pynvml}. 
$M_{\mathrm{weights}}$ represents the allocated VRAM after the model is loaded, while $M_{\mathrm{act}}$ is the additional VRAM observed at the onset of generation. 
Since both reference measurements are derived only from the training split (\S\ref{sec:experiments-dataset}), the test-set evaluations below involve no data leakage.
Table~\ref{tab:primary-results} reports the resulting forecasting metrics for all baselines.

\begin{table*}[t]
	\centering
	\caption{Test-set peak VRAM forecast metrics across the four evaluated LLM backbones}
	\label{tab:primary-results}
	\begin{tabular}{llrrrr}
		\toprule
		Agent LLM & Method & MAPE (\%) & MAE (MiB) & Undercov.\ rate & OAW (\% ceil.) \\
		\midrule
		mistral-7b-instruct-v0.3 & B0 & 10.84 & 789.3 & 0.000 & 0.96 \\
		mistral-7b-instruct-v0.3 & B1 & 7.09 & 512.7 & 0.067 & 1.76 \\
		mistral-7b-instruct-v0.3 & B2 & 6.52 & 458.6 & 0.067 & 1.56 \\
		mistral-7b-instruct-v0.3 & B3 & 7.07 & 510.6 & 0.067 & 1.77 \\
		mistral-7b-instruct-v0.3 & B5 & 4.44 & 333.0 & 0.373 & 0.26 \\
		phi-4-mini & B0 & 14.96 & 846.7 & 0.000 & 1.03 \\
		phi-4-mini & B1 & 0.15 & 8.3 & 0.000 & 0.06 \\
		phi-4-mini & B2 & 0.24 & 13.8 & 0.022 & 0.05 \\
		phi-4-mini & B3 & 0.16 & 8.7 & 0.000 & 0.06 \\
		phi-4-mini & B5 & 2.58 & 146.2 & 0.222 & 0.15 \\
		qwen2.5-coder-14b & B0 & 10.67 & 1280.4 & 0.000 & 1.56 \\
		qwen2.5-coder-14b & B1 & 4.71 & 566.3 & 0.067 & 2.16 \\
		qwen2.5-coder-14b & B2 & 4.41 & 521.0 & 0.067 & 1.92 \\
		qwen2.5-coder-14b & B3 & 4.71 & 565.4 & 0.067 & 2.16 \\
		qwen2.5-coder-14b & B5 & 2.44 & 304.3 & 0.267 & 0.19 \\
		qwen2.5-coder-7b & B0 & 4.85 & 328.2 & 0.067 & 0.39 \\
		qwen2.5-coder-7b & B1 & 3.70 & 252.3 & 0.067 & 0.92 \\
		qwen2.5-coder-7b & B2 & 3.43 & 231.3 & 0.067 & 0.82 \\
		qwen2.5-coder-7b & B3 & 3.70 & 252.1 & 0.067 & 0.92 \\
		qwen2.5-coder-7b & B5 & 2.20 & 153.5 & 0.573 & 0.07 \\ 
		\bottomrule
	\end{tabular}
\end{table*}

Before the analysis, we must contextualize the metrics reported in Table~\ref{tab:primary-results}. 
First, the \ac{oaw} metric divides the mean slack by the full 80\,GB H100 capacity ceiling; consequently, the reported percentages appear small even when the absolute slack in MiB is substantial. 
In no instance does this represent an observed hardware \ac{oom} event or an admission-control outcome at a specific GPU capacity. 
Second, to ensure reproducibility of these results, the random initialization and shuffled training order of B2's frozen-encoder head are pinned via a fixed random seed (\texttt{B2\_RANDOM\_SEED}, detailed in \S\ref{sec:appendix-reproduce-table2}). 

Finally, the undercoverage rates cannot be read as a single ranked column across all rows. 
For the empirical models (B1--B3), the undercoverage rate is calculated against a conservative $z_\alpha$-inflated interval ($\Pr(M_{\mathrm{peak,true}} > M_{\mathrm{upper,pred}})$ with $z_\alpha=1.96$ and $\hat\sigma$ as the train-split residual standard deviation). 
In contrast, B0 and B5 lack a fitted residual term; their undercoverage rate strictly evaluates the bare point prediction ($\Pr(M_{\mathrm{peak,true}} > M_{\mathrm{peak,pred}})$). 
Since B5 tracks the mean of the true distribution rather than a conservative upper bound, its naturally higher undercoverage rate (0.22--0.57) reflects a well-centered point estimate, not a calibration failure, and is not directly comparable to the interval-based B1--B3 statistics.

With this context established, Table~\ref{tab:primary-results} demonstrates that B0's point prediction (the worst-case bound) undershoots the true peak on only 0--6.7\,\% of test rows, yielding a \ac{mape} of 4.85--14.96\,\%. 
B5, supplied with true step counts and reasoning-expansion trajectories, achieves a highly competitive MAPE of 2.20--4.44\,\%. 
Notably, B5 achieves a lower \ac{mape} than B2 on three of the four backbones: Mistral-7B (4.44\,\% vs.\ 6.52\,\%), Qwen2.5-Coder-7B (2.20\,\% vs.\ 3.43\,\%), and Qwen2.5-Coder-14B (2.44\,\% vs.\ 4.41\,\%). 
When pooled across all 270 test rows, a paired sign test on the absolute percentage error ($|\mathrm{APE}|$) reveals no significant difference between B2 and B5 ($p=0.76$, with B2 winning 132 rows and B5 winning 138).

Crucially, this parity disappears if the closed-form model is improperly calibrated. 
If B0 and B5 are fed the raw GGUF file size instead of the measured $M_{\mathrm{weights}}$, and zero instead of $M_{\mathrm{act}}$, their performance degrades severely. 
\ac{mape} spikes to 16.66--53.25\,\%, and the undercoverage rate hits 100\,\% across every backbone (meaning every single test row is underestimated). 
The accuracy stems entirely from these two measured GPU constants, not just the additive functional form itself.
Phi-4-mini remains the single backbone where B2 decisively outperforms B5 (0.24\,\% vs.\ 2.58\,\% MAPE, $p=9.3\times10^{-9}$, with 41 out of 45 rows favoring B2). 
This divergence is driven by the low variation of Phi-4-mini's true peak VRAM. 
As shown in Table~\ref{tab:mpeak-distribution-test}, its test-split \ac{cv} is 0.32\,\% (the lowest) meaning virtually every trajectory makes the same memory footprint. 
Both B3 (constant mean) and B2 (regression) effectively predict this static value and win by default. 
Since B5 scales with actual step counts and context growth, it generates dynamic predictions for a variable that barely moves. 
This variance mismatch, not any shortcoming in the formula, drives B5's residual error on this specific backbone.
Ultimately, the closed-form model is not broken for agentic workloads. 
Provided the static inputs are marked with correct hardware readings, it matches or outperforms the best learned model on three of the four backbones, losing only when the underlying memory usage lacks enough variance. 

\subsection{Retry-Count Distribution}
\label{sec:results-retry}

During initial pilot experiments, we observed that restricting the agent to a standard \texttt{retry\_budget} of 3 artificially truncated the interaction loop, forcing a bimodal distribution where a significant fraction of trajectories were prematurely terminated at the ceiling. 
To prevent this and capture the agent's natural convergence behavior, we expanded the budget to \texttt{retry\_budget}{=}6 for the final evaluated corpus. 

\begin{table}[h]
	\centering
	\caption{Global retry-count histogram: pilot vs. final}
	\label{tab:retry-histogram}
	\begin{tabular}{lrrrr}
		\toprule
		& $r{=}0$ & $r{=}1$ & $r{=}2$ & $r{=}3$ \\
		\midrule
		Pilot ($n{=}1800$) & 1054 & 50 & 24 & 672 \\
		Final ($n{=}1920$) & 579 & 399 & 174 & 768 \\
		\bottomrule
	\end{tabular}
\end{table}

Table~\ref{tab:retry-histogram} reports the resulting retry-count distribution on our published dataset ($n=1920$). 
Under the expanded budget, the probability mass distributes naturally across intermediate counts. 
Although the empirical maximum observed retry count in the final corpus remains 3, the expanded ceiling ensures the censoring rate is exactly 0\,\%. 
The distribution therefore reflects the true halting behavior of the agentic loop rather than an enforced cutoff.

\subsection{\texorpdfstring{$N_{\mathrm{true}}$}{N\_true} Tail Characterization}
\label{sec:results-ntrue}

Table~\ref{tab:ntrue-summary} summarizes $N_{\mathrm{true}}$ (total tokens across all node executions) per backbone across the full evaluated corpus ($n=1920$). 
The global distribution is heavy-tailed: while the modal bin is just $N_{\mathrm{true}}{=}6$ (comprising 153 of 1920 rows), individual runs span four orders of magnitude, reaching a maximum of 29,488 tokens.

\begin{table}[h]
	\centering
	\caption{$N_{\mathrm{true}}$ summary per LLM backbone ($n=1920$)}
	\label{tab:ntrue-summary}
	\begin{tabular}{lrrrr}
		\toprule
		Backbone LLM & $N{=}0$ & mean & median & max \\
		\midrule
		Phi-4-mini & 0/300 & 2301 & 2009 & 9{,}812 \\
		Mistral-7B-Instruct & 0/540 & 2185 & 1050 & 7{,}704 \\
		Qwen2.5-Coder-7B & 1/540 & 1929 & 980 & 29{,}488 \\
		Qwen2.5-Coder-14B & 0/540 & 1315 & 1172 & 5{,}844 \\
		\midrule
		\textbf{Global} & \textbf{1/1920 (0.05\,\%)} & --- & --- & --- \\
		\bottomrule
	\end{tabular}
\end{table}

During preliminary testing, we observed that standard prompt configurations frequently allowed the agent to short-circuit before entering the tool loop, resulting in a high rate of empty trajectories ($N_{\mathrm{true}}{=}0$). 
By combining an expanded \texttt{retry\_budget}{=}6 with adversarial, retry-provoking prompts in our final corpus, we effectively suppressed this failure mode, dropping the $N_{\mathrm{true}}{=}0$ rate to just 0.05\,\% (1 out of 1920 runs). 

Finally, the overdispersion (the variance-to-mean ratio) of $N_{\mathrm{true}}$ on the training split lands high at 3,033. 
\textit{This severe overdispersion empirically demonstrates that modeling token generation as a standard Poisson process is incorrect, invalidating the use of a simple Poisson head for token-count prediction in this regime}.

\subsection{Compile-Success Rates per Backbone and Prompt Slice (F2)}
\label{sec:results-compile}

Table~\ref{tab:compile-slice} reports the corpus-wide \texttt{compile\_success} rates across the different LLM backbones and prompt categories. 
These figures are aggregated from a highly granular prompt-family breakdown encompassing all 1,920 evaluated trajectories.

\begin{table}[h]
	\centering
	\caption{Corpus-wide \texttt{compile\_success} rate (\%) per LLM backbone and prompt category}
	\label{tab:compile-slice}
	\begin{tabularx}{\linewidth}{Xrrr}
		\toprule
		Backbone LLM & KernelBench & clean\_synthetic & retry\_provoking \\ 
		\midrule
		Phi-4-mini & 4.0 & 7.0 & 6.0 \\
		Mistral-7B-Instruct & 4.8 & 11.1 & 14.7 \\
		Qwen2.5-Coder-7B & 28.9 & 40.0 & 40.2 \\
		Qwen2.5-Coder-14B & 53.6 & 64.2 & 67.4 \\
		\bottomrule
	\end{tabularx}
\end{table}

As demonstrated in Table~\ref{tab:compile-slice}, corpus-wide compile success strictly bifurcates by model capacity and specialization. 
The smaller or generalist models struggle (Phi-4-mini averaging 5.7\,\%, Mistral-7B averaging 10.4\,\%), whereas the larger, coding-specialized models succeed at a substantially higher rate (Qwen2.5-Coder-7B at 36.7\,\%, Qwen2.5-Coder-14B at 62.0\,\%). 
Crucially, this capacity-driven bifurcation remains consistent across all three prompt categories (see \S\ref{sec:limitations} for further disambiguation between the test split and the full corpus).

\begin{figure}[htpb]
	\centering
	\includegraphics[width=\linewidth]{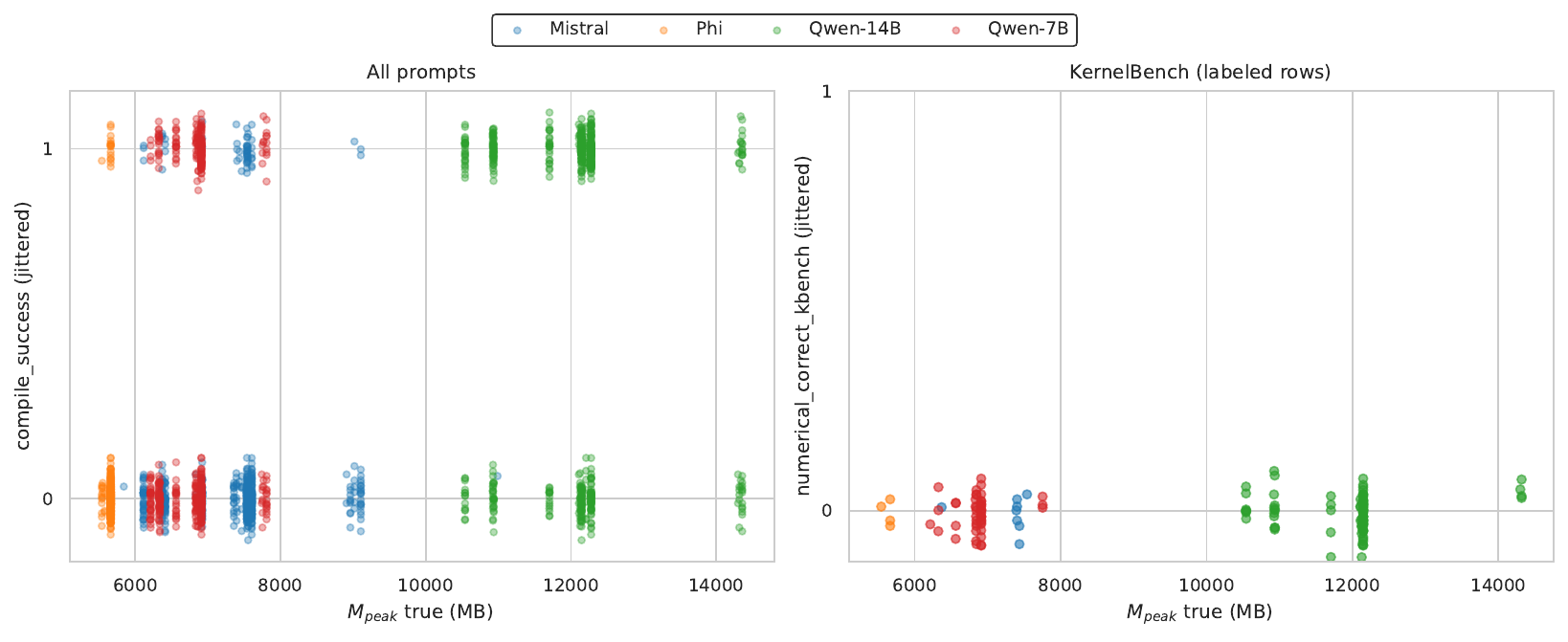}
	\caption{Peak VRAM versus correctness per LLM backbone across the evaluated corpus. Left: $M_{\mathrm{peak,true}}$ vs.\ \texttt{compile\_success} (jittered) for all 1,920 traces. Right: $M_{\mathrm{peak,true}}$ vs.\ \texttt{numerical\_correct\_kbench} (jittered) for the 149 KernelBench rows with a defined label.}
	\label{fig:vram-correctness-joint}
\end{figure}

Figure~\ref{fig:vram-correctness-joint} examines whether \texttt{compile\_success} is confounded by peak VRAM consumption.
Within each backbone, the point-biserial correlation between $M_{\mathrm{peak,true}}$ and \texttt{compile\_success} is small and statistically insignificant for three of the four models (Mistral $r=0.012$, $p=0.79$; Phi-4-mini $r=-0.004$, $p=0.94$; Qwen2.5-Coder-14B $r=-0.065$, $p=0.13$). 
Qwen2.5-Coder-7B exhibits a minor but statistically significant positive correlation ($r=0.133$, $p=0.0019$, $n=540$), indicating that successful compilations skew marginally higher in peak VRAM. 
Conversely, for \texttt{numerical\_correct\_kbench} (Figure~\ref{fig:vram-correctness-joint}, right panel), no correlation is computable: every one of the 149 evaluated KernelBench trajectories is labeled \texttt{False}, for the four harness reasons detailed next in \S\ref{sec:results-numerical-f3}. 

\subsection{Numerical Correctness and Empirical Baselines (F3)}
\label{sec:results-numerical-f3}

While the compile-success bifurcation (F2) clearly demonstrates that functional code synthesis depends on LLM backbone capacity, evaluating full numerical correctness on KernelBench proved impossible at scale due to tracing-harness limitations. 
KernelBench requires specific PyTorch extension scaffolding (\texttt{load\_inline}) that AgentK does not naturally elicit. 

We attempted to evaluate the 149 successfully compiled kernels using a synthesized, signature-agnostic wrapper, but every single row failed before the actual numerical check (\texttt{torch.allclose}) could execute. 
The failures were entirely systemic: 116 failed during the wrapper's internal JIT build (with exact errors lost to log truncation), 28 failed a pre-flight \texttt{torch.load} version check, 4 exceeded the wrapper's argument limits, and 1 timed out. 
Since no kernel actually reached the numerical evaluation stage, the \texttt{numerical\_correct\_kbench=False} labels in our dataset reflect an inadequate testing harness, not flawed LLM math. 
Consequently, we rely exclusively on \texttt{compile\_success} as our correctness signal and defer signature-aware testing to future work.

Returning to VRAM forecasting (Table~\ref{tab:primary-results}), we assess whether learned regression (B2) justifies its complexity over a simple constant mean (B3). 
On three of the four backbones (Mistral-7B, Qwen2.5-Coder-7B, and Qwen2.5-Coder-14B), B2 marginally outperforms B3, reducing the aggregate \ac{mape} by a modest 6--8\,\%. 
However, these improvements carry overlapping 95\,\% confidence intervals. 

On the lowest-variance backbone (Phi-4-mini), B2 performs measurably worse than B3, increasing relative error by 57\,\% (0.24\,\% vs.\ 0.16\,\% MAPE). 
Since Phi-4-mini's true VRAM usage is nearly static (Table~\ref{tab:mpeak-distribution-test}), the regression model ends up fitting noise rather than meaningful variance. 
While B2 wins the majority of individual test rows on the three larger backbones, B3 decisively wins on Phi-4-mini. 
Ultimately, in a regime so heavily dominated by static weights, deploying a complex empirical predictor like B2 offers negligible operational benefit over a simple historical average.

\subsection{Variance Decomposition}
\label{sec:results-variance}

We use a one-way analysis of variance (ANOVA) to quantify how specific categorical factors drive peak VRAM variation. 
Table~\ref{tab:variance-decomposition} partitions the total variance in our dataset into the proportion explained by prompt family, retry count, and hardware sweep configuration, versus the unexplained residual noise. \vspace{-1em}

\begin{table*}[htpb]
	\centering
	\caption{One-way ANOVA of $M_{\mathrm{peak,true}}$ (MiB) by grouping factor}
	\label{tab:variance-decomposition}
	\begin{tabular}{llrrrrr}
		\toprule
		Backbone LLM & Grouping & $\eta^2$ & $F$ & $p$ & $n$ & Residual frac. \\
		\midrule
		Mistral & Prompt family & 0.000 & 0.02 & 9.81e-01 & 540 & 1.000 \\
		Mistral & Retry count & 0.001 & 0.23 & 8.77e-01 & 540 & 0.999 \\
		Mistral & Config sweep & 0.913 & 493.97 & 3.83e-122 & 240 & 0.087 \\
		Phi & Prompt family & 0.126 & 21.34 & 2.18e-09 & 300 & 0.874 \\
		Phi & Retry count & 0.002 & 0.21 & 8.90e-01 & 300 & 0.998 \\
		Phi & Config sweep & --- & --- & --- & 0 & --- \\
		Qwen-14B & Prompt family & 0.001 & 0.14 & 8.71e-01 & 540 & 0.999 \\
		Qwen-14B & Retry count & 0.004 & 0.66 & 5.78e-01 & 540 & 0.996 \\
		Qwen-14B & Config sweep & 1.000 & 831324.07 & 0.00e+00 & 240 & 0.000 \\
		Qwen-7B & Prompt family & 0.000 & 0.02 & 9.82e-01 & 540 & 1.000 \\
		Qwen-7B & Retry count & 0.056 & 10.60 & 8.82e-07 & 540 & 0.944 \\
		Qwen-7B & Config sweep & 0.999 & 92491.60 & 0.00e+00 & 240 & 0.001 \\
		\bottomrule
	\end{tabular}
\end{table*}

Across all four backbones, prompt family and retry count explain at most 12.6\,\% of the total variance, leaving 87.4\,\% to 100\,\% as unexplained residual noise (excluding the deliberately perturbed configuration sweep rows). 
Even where retry count shows a statistically significant effect (Qwen2.5-Coder-7B, $p=8.8\times10^{-7}$), it explains less than 6\,\% of that backbone's variance ($\eta^2=0.056$). Residual histograms for these cells are provided in Appendix~\ref{sec:appendix-variance}.
This lack of variance confirms a strictly \emph{weight-term-dominated} memory regime. 
Under Q4\_K\_M quantization, static model weights ($M_{\mathrm{weights}}$) establish a massive, near-constant memory floor. 
Any dynamic, per-run fluctuations originating from the expanding KV-cache ($M_{KV}(N)$) or temporary activations ($M_{\mathrm{act}}$) are trivially small relative to this fixed weight term.

Consequently, a simple constant-mean baseline (B3) performs exceptionally well because it directly captures this massive static floor.
However, this lack of variance does not invalidate the closed-form model (B5). 
As demonstrated in \S\ref{sec:results-mpeak-f1}, once properly calibrated with live GPU measurements for $M_{\mathrm{weights}}$ and $M_{\mathrm{act}}$, the analytical formula accurately anchors to this same dominant floor while making the necessary, albeit tiny, dynamic adjustments. 
This structural soundness is precisely why B5 matches or outperforms the learned regression model (B2) on three of the four evaluated backbones.

%% file: sections/discussion.tex
\section{Discussion}
\label{sec:discussion}

Our central finding is structural: when given hindsight on step count and reasoning expansion $(N, E)$ alongside live-measured constants ($M_{\mathrm{weights}}$ and $M_{\mathrm{act}}$), the closed-form model (B5) matches or beats the best learned baseline (B2) on three of four backbones. 
This proves the additive memory decomposition itself is sound. 
However, because $(N, E)$ are only known after a run finishes, B5 cannot be deployed directly at admission time without a separate upstream predictor for those trajectory parameters.

The model's accuracy stems from weight dominance. 
Under Q4\_K\_M, static weights form a massive memory floor. 
Dynamic runtime changes (KV-cache and activations) are so small that minor formula errors have negligible impact. 
This extreme stability also explains why a simple constant-mean baseline (B3) performs competitively; with prompt features and retries explaining less than 12.6,\% of total variance, there is simply minimal variance left for any model to capture.
Phi-4-mini highlights the boundary of this approach. 
Since its true peak VRAM has near-zero variance (CV 0.32,\%), B5's sensitivity to step counts introduces unnecessary error on a flat target, allowing static predictors (B2, B3) to prevail.
A critical practical lesson is that closed-form models are extremely sensitive to calibration. Feeding raw on-disk file sizes instead of live GPU measurements causes catastrophic undercoverage (100,\% underestimation) because weight dominance leaves no room for input errors.

Ultimately, practical deployment choice depends on timeline. 
At admission time—before an agent executes—engineers must choose among a safe worst-case bound (B0), a constant mean (B3), or prompt-feature regression (B2). 
Using B5 requires a separate proxy to predict $(N, E)$ beforehand. 
Our results establish that if such a proxy is available, the closed-form accounting step will not be the system bottleneck.

%% file: sections/limitations.tex
\section{Limitations}
\label{sec:limitations}


\textbf{Single weight-quantization family.} All evaluations utilize the Q4\_K\_M quantization scheme via \texttt{llama.cpp}. We do not evaluate alternative quantization regimes or precisions (e.g., Q8\_0, FP16, AWQ, or GPTQ variants).

\textbf{Stratified split.} The deterministic split seed disproportionately allocated compile-easier prompts to the test slice (e.g., the Qwen2.5-Coder-14B compile-success rate is 78.7\,\% on the test split versus 62.0\,\% across the full corpus). We deliberately did not resample in order to rigorously preserve pre-registered split determinism; test-set MAPE claims remain unaffected because MAPE conditions directly on $M_{\mathrm{peak}}$, not on \texttt{compile\_success}.

\textbf{Numerical correctness.} Full numerical correctness on KernelBench was not measurable at scale (\S\ref{sec:results-numerical-f3}): 100\,\% of the \texttt{nvcc}-compile-success trajectories lacked native \texttt{ModelNew}/\texttt{load\_inline} scaffolding, requiring a synthesized wrapper. 
All the 149 wrapped rows were blocked before \texttt{torch.allclose} could execute; 116 failed due to a JIT-build error inside the wrapper, 28 due to an unrelated \texttt{torch.load} version check, 4 by exceeding the wrapper's arity limit, and 1 by a genuine timeout. This represents a strict limitation of our tracing harness format, not an empirical finding regarding the underlying model capabilities.

\textbf{Closed-form trajectory correlation.} While a Pearson correlation of $r = 0.21$ was measurable on an earlier preliminary pilot batch ($n = 846$), recomputation on the final evaluated corpus is mathematically undefined for 94\,\% of the traces. This is due to the near-constant boundary VRAM floor imposed by the Q4\_K\_M quantization regime.

\textbf{B5 is partially, not fully, oracle-fed.} Our oracle baseline (B5, \S\ref{sec:experiments-baselines}) provides the true node count $N$ and true non-tool completion-token rate $E$, and recovers $L_{\mathrm{base}}$ from the run's own runtime tokenizer. 
However, it does \emph{not} oracle-feed the tool-return term. 
Since our trace records only per-node average completion tokens (an LLM-generated quantity) rather than external tool-return content (e.g., retrieved-document) as a separate field, no ground-truth tool-return total exists in this dataset to oracle-feed. 
We therefore set this term to zero to avoid double-counting against $E$. 
Therefore, B5 already matches or beats the best learned baseline on three of four backbones, making this a conservative choice rather than one that artificially inflates B5's reported accuracy.

\textbf{Weight/activation calibration relies on a single reference run per backbone.} The $M_{\mathrm{weights}}$ and $M_{\mathrm{act}}$ constants used by B0 and B5 (\S\ref{sec:results-mpeak-f1}) are each measured from a single live GPU reading per backbone (per key/value cache-type pair), rather than averaged across multiple runs. We confirmed that every reference run belongs exclusively to the training split, ensuring zero test-set leakage; however, using a single measurement per backbone means we cannot report statistical confidence intervals on these two constants.

%% file: sections/app_abbreviations.tex
\section{List of Abbreviations}
\label{sec:app_abbrev}

\vspace{1em}

\acsetup{list/template = tabular} 

\printacronyms[heading=none]

%% file: appendices/appendix_archoverview.tex
\section{AgentK: Architecture Overview}
\label{sec:appendix-agentk}

Although AgentK is an open source LangGraph based agent which has been available for a long time, since it was deveolped by the authors, for the anonymity purposes the actual name and repository link could not be added in this paper.
Instead, this appendix summarizes AgentK's control flow at the level of node roles and routing decisions. 
This provides a concrete picture of the tracing unit referenced throughout \S\ref{sec:method} and \S\ref{sec:experiments} (specifically concerning node executions $N$, the tool-call term $P(T_i)$, and the \texttt{retry\_budget}) without disclosing implementation-identifying details. 
The underlying prompts, the reference-documentation corpus, and the model-routing configurations are provided in the anonymized code artifact linked in \S\ref{sec:method}.

\subsection{Pipeline Stages}
\label{sec:appendix-agentk-stages}

AgentK operates as a five-node cyclic graph rather than a single-pass pipeline. Four of the nodes execute at most once per task, while one specific edge (Critic $\to$ Generator) forms a deliberate feedback loop bounded by the \texttt{retry\_budget}.

\textbf{Planner.} Classifies the incoming prompt as in-scope (a kernel-synthesis task) or out-of-scope. It short-circuits out-of-scope prompts before any generation cost is incurred, emitting a schema-constrained boolean routing decision.

\textbf{Analyzer.} Retrieves the top-$k$ passages from a locally indexed reference-documentation corpus and produces a natural-language technical analysis of the task (e.g., memory-access patterns, parallelism opportunities, and expected bottlenecks). This retrieval step represents the tool invocation $T_i$ referenced in \Eqref{eq:Lt}.

\textbf{Optimizer.} Converts the Analyzer's free-form analysis into a schema-constrained optimization strategy (e.g., launch-configuration and memory-placement fields), ensuring that the downstream generation consumes strictly typed fields rather than unconstrained free text.

\textbf{Generator.} Synthesizes the target source code implementing the strategy. During a retry iteration, it additionally receives the previously generated (failing) source code alongside the exact compiler diagnostic from the Critic, and is explicitly instructed to address every reported error.

\textbf{Critic.} Invokes the reference compiler as a ground-truth verification gate. A successful compilation terminates the loop and records a success trace. A failure either (a) feeds the diagnostic back to the Generator, provided the retry budget is not yet exhausted, or (b) terminates the loop and records a failure trace once the budget is depleted. This specific node acts as the source for the \texttt{compile\_success} label utilized throughout \S\ref{sec:results}.

\begin{figure}[htpb]
	\centering
	\includegraphics[width=\linewidth]{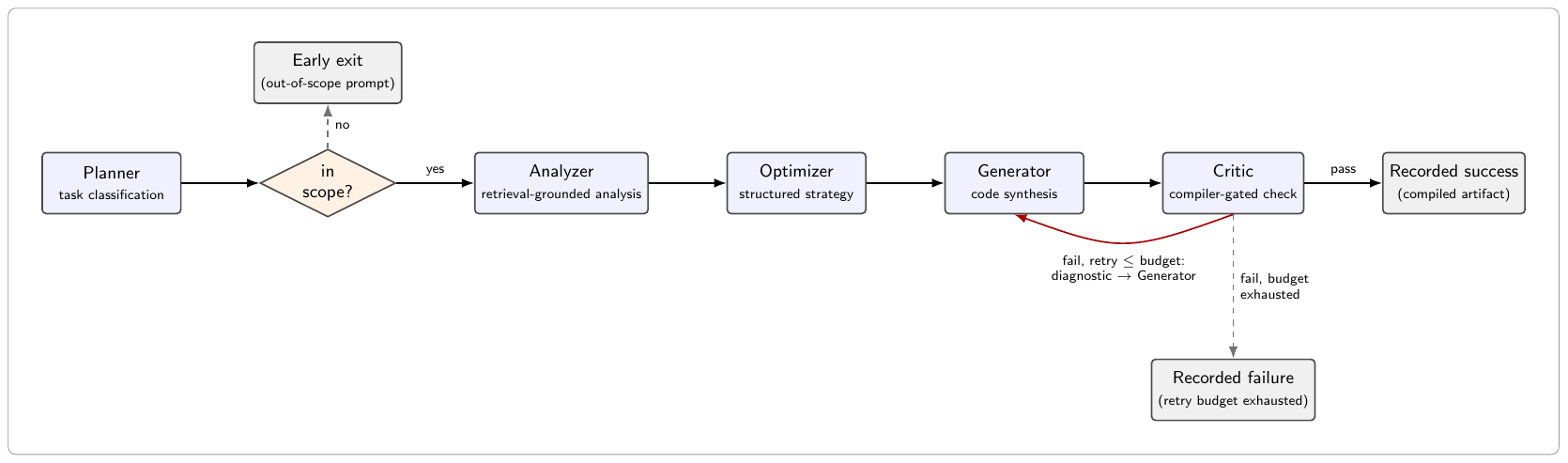}
	\caption{AgentK end-to-end control flow. Solid arrows represent the forward pipeline; the dashed arrow denotes the out-of-scope short-circuit exit; the red arrow highlights the compiler-diagnostic feedback edge, taken on every compile failure until the \texttt{retry\_budget} is reached. Each traversal of the Generator or Critic constitutes one node execution contributing to $N$ and to $N_{\mathrm{true}}$ (\S\ref{sec:method}).}
	\label{fig:agentk-flow}
\end{figure}

\subsection{Retry Loop and Trace Boundaries}
\label{sec:appendix-agentk-retry}

The Planner, Analyzer, and Optimizer nodes each execute a maximum of once per task. The Generator and Critic form the compile-repair loop and may each execute up to the \texttt{retry\_budget} limit (configured to 6 for the final evaluated corpus; \S\ref{sec:results-retry}). Every node execution---including repeated Generator and Critic rounds within a single task---contributes completion tokens to the traced $N_{\mathrm{true}}$ measure (defined as the total completion tokens across all node executions; see Appendix~\ref{sec:appendix-datasheet} for exact field semantics) and adds one unit to the node-execution count $N$ used in the closed-form decomposition in \S\ref{sec:method-formulation}.

Consequently, a task that fails on every attempt up to the budget, and a task that succeeds on its very first attempt, differ substantially in both $N$ and $N_{\mathrm{true}}$. This structural variance forms the fundamental mechanism underlying the retry-count and $N_{\mathrm{true}}$ distributions reported in \S\ref{sec:results-retry} and \S\ref{sec:results-ntrue}.

%% file: appendices/appendix_variance.tex
\section{Variance Decomposition of $M_{\mathrm{peak,true}}$}
\label{sec:appendix-variance}

This appendix provides supplementary visualizations and detailed interpretations for the variance decomposition analysis introduced in \S\ref{sec:results-variance}. The full one-way ANOVA table---reporting per-backbone $\eta^2$, $F$, $p$, $n$, and residual fractions for each grouping factor---is available in Table~\ref{tab:variance-decomposition} of the main text.

For context, the ANOVA tests three specific grouping factors:
\begin{itemize}
	\item \textbf{Prompt family}: Evaluated across three distinct prompt sources.
	\item \textbf{Retry count}: The discrete number of compile attempts (matching the \texttt{retry\_count} field summarized in Table~\ref{tab:retry-histogram}).
	\item \textbf{Configuration sweep}: The combined hardware sweep settings $(n_{\mathrm{ctx}}, \texttt{cache\_type\_k}, \texttt{cache\_type\_v})$. This is evaluated exclusively on the \texttt{is\_sweep=True} subsets ($n=240$ per backbone; Phi-4-mini is excluded as it contains no sweep rows).
\end{itemize}

\begin{figure}[htpb]
	\centering
	\includegraphics[width=\linewidth]{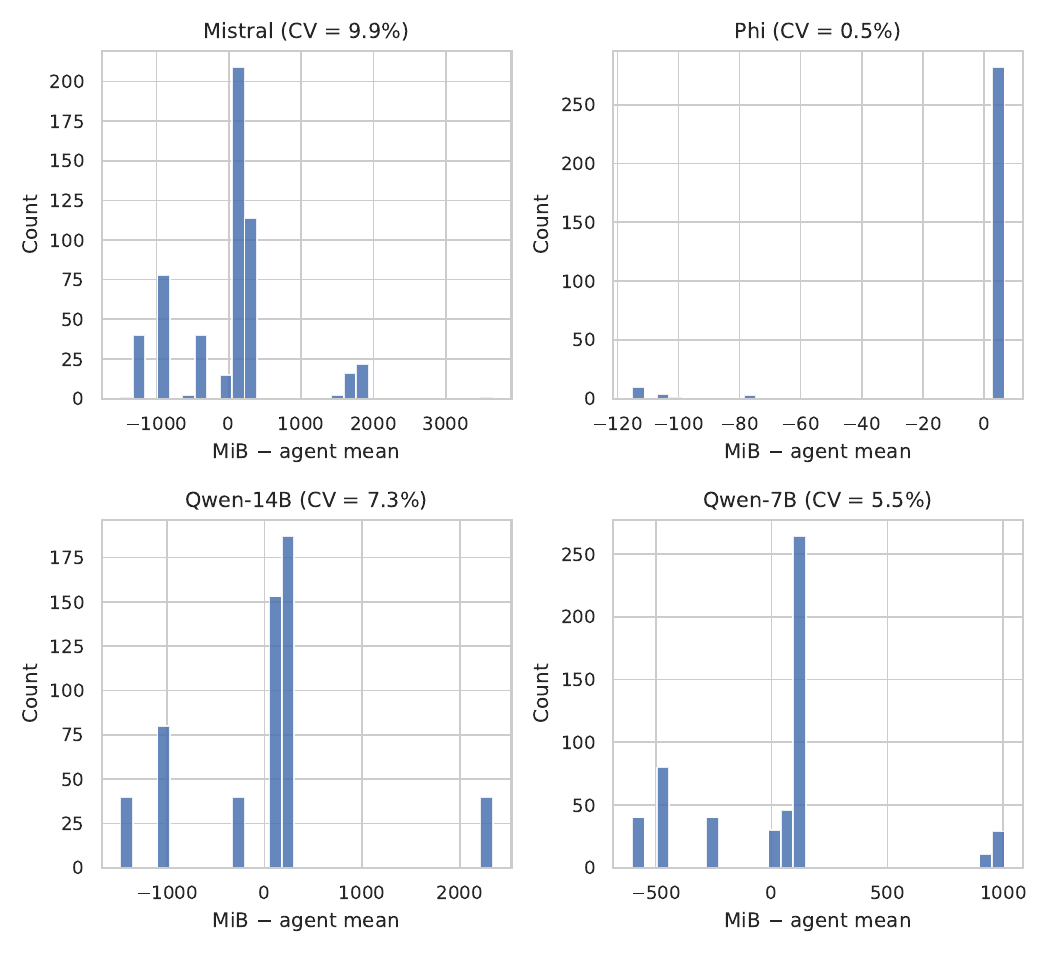}
	\caption{Per-backbone histograms of $M_{\mathrm{peak,true}} - \bar{M}_{\mathrm{peak}}$ (MiB) across the full evaluated corpus. The narrow residual spreads under Q4\_K\_M are visible for Phi-4-mini and Qwen2.5-Coder-7B; Mistral-7B and Qwen2.5-Coder-14B exhibit wider tails driven specifically by the $n_{\mathrm{ctx}}$ and KV-cache hardware sweep rows.}
	\label{fig:variance-residuals}
\end{figure}

\subsection*{Interpretation of Results}

\textbf{Minimal Variance from Prompts and Retries.} Across the entire per-backbone corpus, prompt family and retry count jointly explain a maximum of only 12.6\,\% of the total $M_{\mathrm{peak,true}}$ variance (observed specifically on the Phi-4-mini prompt family, where $\eta^2=0.126$). In all other cases, these factors yield an $\eta^2 \leq 5.6\,\%$ (peaking with the Qwen-7B retry count at $\eta^2=0.056$). This leaves an overwhelming 87\,\% or more as unexplained residual variance in every standard evaluation cell.

\textbf{Strict Weight-Term Dominance.} This lack of variance reinforces the reality of a strictly weight-term-dominated memory regime under Q4\_K\_M quantization. The static model weights ($M_{\mathrm{weights}}$) establish a massive, nearly immovable baseline memory footprint for each backbone. Dynamic, per-run fluctuations caused by the expanding KV-cache ($M_{KV}(N)$) or temporary activations ($M_{\mathrm{act}}$) are trivially small relative to this baseline.

\textbf{Configuration Sweeps as the Exception.} The configuration-sweep rows operate as an exception by design. On the \texttt{is\_sweep=True} subsets, the architectural configurations $(n_{\mathrm{ctx}}, \texttt{cache\_type\_k/v})$ explain virtually all the variance ($\eta^2 \approx 0.91$--$1.00$). This confirms that KV-cache mass does scale predictably with $n_{\mathrm{ctx}}$ when weights are held fixed. However, this axis of variance is artificially induced and entirely orthogonal to the natural task-level variation seen in the standard evaluated corpus.

\textbf{Implications for Baseline Forecasting.} Ultimately, this variance decomposition explains why a simple constant-mean baseline performs so effectively. When non-weight terms contribute only a few percent to the total variance during typical runs, predicting the training-set mean captures nearly all available signal. However, this does not invalidate the closed-form model. As demonstrated in \S\ref{sec:results-mpeak-f1}, once $M_{\mathrm{weights}}$ and $M_{\mathrm{act}}$ are anchored to live GPU measurements, the closed-form model successfully tracks this same dominant weight term while applying the correct (albeit tiny) dynamic adjustments, allowing it to match or beat the best learned baseline on three of the four evaluated backbones.

%% file: appendices/appendix_dataset_v2.tex
\section{Datasheet: AgentK Evaluation Corpus}
\label{sec:appendix-datasheet}

\paragraph{Motivation.}
This dataset was created to characterize the peak-VRAM behavior of quantized-LLM agentic workloads for the measurement study reported in this paper. 
It is not a general-purpose code-generation or systems benchmark; rather, each row represents one instrumented AgentK execution under strictly controlled hardware and quantization settings, complete with per-run VRAM telemetry and compile-outcome labels.

\paragraph{Composition.}
Each instance represents a single traced trajectory: a tuple of the LLM backbone, prompt, inference configuration, and hardware sweep flag. 
The release contains exactly 1,920 runs across four LLM backbones (Phi-4-mini, Mistral-7B-Instruct-v0.3, Qwen2.5-Coder-7B, and Qwen2.5-Coder-14B) and 300 unique CUDA-kernel-synthesis prompts. 
The canonical row index and per-slice compile-success aggregations are provided in the \texttt{data/} directory of the accompanying reproducibility artifact.

\paragraph{Collection process.}
Data were acquired using the AgentK tracing harness described in \S\ref{sec:method} and the anonymous reproducibility artifact cited in \S\ref{sec:introduction}. 
Data collection was performed on a single NVIDIA H100 80\,GB GPU (with Multi-Instance GPU / MIG disabled), running Python 3.10.14 and precisely pinned dependencies (\texttt{torch==2.2.2}, \texttt{llama-cpp-python==0.3.33}, \texttt{langgraph==1.2.9}). 
KernelBench reference problems are pinned to upstream commit \texttt{423217d9fda91e0c2d67e4a43bf62f96f6d104f1}. 
Trace collection was completed in late July 2026 across two phases: a base configuration trace and a subsequent context/KV-cache hardware sweep.

\paragraph{Preprocessing, cleaning, and labeling.}
Downstream metrics are aggregated and labeled via the artifact's build scripts. 
Key metric definitions include: \texttt{N\_true}, which represents the total completion tokens summed across all node events in the trajectory (not simply the node count), and \texttt{N\_nodes\_true}, which retains the discrete node-execution count for closed-form trajectory indexing. 
For the supplementary counterfactual admission-control analysis --- not an observed hardware out-of-memory event, and not the 80\,GB H100 ceiling used elsewhere for \ac{oaw} normalization --- evaluations apply a fixed, synthetic VRAM budget grid $B \in \{2,4,6,8,10,12,16,24,40\}$\,GB and determine mathematically whether $M_{\mathrm{peak,true}} > B$ at each hypothetical capacity.

\paragraph{Uses.}
To date, this dataset has been utilized strictly for the VRAM-forecasting and agent-workload characterization analyses presented in this paper. 
It is suitable for follow-up research on VRAM forecasting, agent-framework telemetry, and quantization sensitivity under fixed agent semantics. 
It is not suitable for evaluating LLM code-generation quality at scale: \texttt{compile\_success} serves only as a coarse proxy, and our numerical KernelBench correctness harness could not get any of the 149 wrapped rows as far as \texttt{torch.allclose} (see the four-way failure breakdown in \S\ref{sec:results-numerical-f3}).

\paragraph{Distribution.}
During the anonymous review period, the dataset manifest, derived/raw dataset archives, and code artifact are securely hosted at \url{https://anonymous.4open.science/r/agentk-tmlr-anonymous/}. 
A public, deanonymized URL will be provided upon acceptance. 
The primary dataset artifacts are released under the \textbf{Apache-2.0} license. 
KernelBench prompts are included under their original upstream MIT license. 
The trace data contains LLM-generated text from the four evaluated backbones, which are subject to their respective model licenses (Apache-2.0 / MIT).

\paragraph{Maintenance.}
The dataset is maintained by the authors. 
During the review phase, contact should be routed via OpenReview; post-acceptance, maintenance and correspondence will transition to the public repository's issue tracker. Versioning follows semantic versioning (SemVer) via git tags on the release commit.

%% file: appendices/appendix_reproduce_table2.tex
\section{Reproducing Table~\ref{tab:primary-results} Without a GPU}
\label{sec:appendix-reproduce-table2}

Table~\ref{tab:primary-results} relies on the B2 baseline (a regression head trained on frozen sentence embeddings), which requires a GPU to train and introduces minor run-to-run randomness via weight initialization and data-loader shuffling. To allow reviewers to verify the table's exact numbers without needing a GPU, retraining models, or network access, we provide the exact test-split predictions used to generate the table. These are shipped as frozen parquet files alongside a standalone script that recalculates all metrics (MAPE, MAE, undercoverage rate, and \ac{oaw}) directly from the released \texttt{data/} directory.

\paragraph{Prediction Artifacts.}
The predictions are stored at

\path{data/frozen_predictions_table2/{agent_llm}/{method}/test.parquet}. 

There is one file for each of the four backbones across all five evaluated methods (B0, B1, B2, B3, B5). Each file contains one row per test-split prompt, featuring two key columns: \texttt{Mpeak\_pred} (the point prediction) and \texttt{Mupper\_pred} (the upper bound used for undercoverage and \ac{oaw} metrics). As established in \S\ref{sec:results-mpeak-f1}, for the analytical models (B0, B5), \texttt{Mupper\_pred} is simply equal to \texttt{Mpeak\_pred}.

\paragraph{How to Reproduce.}
Reviewers can regenerate the table metrics from the root directory of the anonymous artifact using the following command:
\begin{quote}
	\texttt{python scripts/reproduce\_table2.py} \\
	\emph{(equivalently:} \texttt{make reproduce-table2}\emph{)}
\end{quote}
This script deterministically re-derives the test-split row order from the provided \texttt{data/labels\_v2/} and \texttt{data/traces\_v2/index.parquet} files. It then merges the corresponding ground-truth $M_{\mathrm{peak,true}}$ with each frozen prediction row, computes the metrics, and outputs both a plain-text summary and the exact \LaTeX{} table body. The script's output matches Table~\ref{tab:primary-results} to the final reported digit.

\paragraph{B2 Determinism.}
The random seed used to initialize B2's linear head and data-loader shuffle is recorded in \path{data/frozen\_predictions\_table2/\{agent\_llm\}/B2/fit\_meta.json}. While retraining B2 end-to-end with this seed on the released data will technically reproduce the predictions, hardware-specific floating-point non-associativity in GPU reductions can cause sub-0.01\,\% drift in the final MAPE on different machines. For exact numerical verification during the review process, we strongly recommend using the provided frozen predictions.

\paragraph{Code Pointers.}
The anonymous code artifact includes a \texttt{REVIEW\_NOTES.md} file that directs reviewers to the source files most relevant to Table~\ref{tab:primary-results}. Key implementations include:
\begin{itemize}
	\item \texttt{src/baselines/b5\_oracle.py}: Enforces the \texttt{Mupper\_pred = Mpeak\_pred} constraint for B5, preventing synthetic interval inflation (\S\ref{sec:results-mpeak-f1}).
	\item \texttt{src/baselines/common.py::accounting\_dict}: Applies the critical live-VRAM calibration constants defined in \texttt{configs/q4\_vram\_calibration\_v2.yaml} (\S\ref{sec:method-q4-accounting}) rather than relying on raw on-disk file sizes.
\end{itemize}